\documentclass[conference]{IEEEtran}
\IEEEoverridecommandlockouts

\usepackage[utf8]{inputenc}
\usepackage[T1]{fontenc}

\usepackage{cite}
\usepackage{amsmath,amssymb,amsfonts}
\usepackage{algorithmic}
\usepackage{graphicx}
\usepackage{textcomp}
\usepackage{xcolor}

\usepackage{tikz}
\usetikzlibrary{shapes.geometric, arrows.meta, positioning}

\usepackage{pgfplots}
\pgfplotsset{compat=1.17}

\usepackage{listings}
\usepackage{caption}

\usepackage{url}

\begin{document}

\title{Intelligent 5S Audit: Application of Artificial Intelligence for Continuous Improvement in the Automotive Industry\\
{\footnotesize \textsuperscript{}}
\thanks{}
}

\author{\IEEEauthorblockN{1\textsuperscript{st} Rafael da Silva Maciel}
\IEEEauthorblockA{\textit{Institute of Science and Technology} \\
\textit{Federal University of São Paulo}\\
São Paulo, Brazil \\
rafael.maciel@unifesp.br}
\and
\IEEEauthorblockN{2\textsuperscript{nd} Lucio Veraldo Jr.}
\IEEEauthorblockA{\textit{Institute of Science and Technology} \\
\textit{Federal University of São Paulo}\\
São Paulo, Brazil \\
lucio.veraldo@infinityacademy3d.com.br}
}

\maketitle

\begin{abstract}
The evolution of the 5S methodology with the support of artificial intelligence techniques represents a significant opportunity to improve industrial organization audits in the automotive chain, making them more objective, efficient and aligned with Industry 4.0 standards. This work developed an automated 5S audit system based on large-scale language models (LLM), capable of assessing the five senses (Seiri, Seiton, Seiso, Seiketsu, Shitsuke) in a standardized way through intelligent image analysis. The system's reliability was validated using Cohen's concordance coefficient ($\kappa = 0.75$), showing strong alignment between the automated assessments and the corresponding human audits. The results indicate that the proposed solution contributes significantly to continuous improvement in automotive manufacturing environments, speeding up the audit process by 50\% of the traditional time and maintaining the consistency of the assessments, with a 99.8\% reduction in operating costs compared to traditional manual audits. The methodology presented establishes a new paradigm for integrating lean systems with emerging AI technologies, offering scalability for implementation in automotive plants of different sizes.
\end{abstract}


\section{Introduction}
The global automotive industry faces growing competitiveness challenges demanding maximized operational efficiency and production quality. The 5S methodology, recognized worldwide as the foundation for workplace organization and cleanliness, plays a strategic role in operational excellence. Originating from Toyota Production System practices in post-war Japan (OHNO, 1997), it comprises five senses - Seiri (use), Seiton (order), Seiso (cleanliness), Seiketsu (standardization) and Shitsuke (discipline) - creating efficient, safe and productive environments through systematic waste elimination.

In automotive manufacturing with minimal error margins and strict quality standards, 5S audits are critical for maintaining lean culture and ensuring continuous improvement (KANABAR et al., 2024). Traditional audits by cross-functional teams face limitations in human subjectivity, resource availability and time requirements for comprehensive facility coverage. These constraints often result in infrequent assessments and delayed identification of non-conformities, impacting overall operational performance.

Simultaneously, industrial automation and artificial intelligence convergence has revolutionized Industry 4.0 process monitoring and optimization. Large Language Models (LLMs) with multimodal capabilities enable contextual visual data analysis with unprecedented precision (LECUN; BENGIO; HINTON, 2015), creating opportunities for automating traditionally human-dependent auditing tasks. This technological advancement promises to address traditional audit limitations while maintaining assessment quality.

Integrating AI systems with lean methodologies represents a promising frontier, offering standardized evaluations, reduced costs and increased monitoring frequency. This paper presents development and validation of an AI-based automated 5S audit system for automotive manufacturing environments.

The main objective demonstrates technical and economic feasibility of automating 5S audits through intelligent image analysis, establishing a new paradigm for integrating lean methodologies with AI technologies. The methodology encompasses theoretical foundation, system development with validation, comparative performance analysis, and operational/economic impact evaluation. Results demonstrate both technical effectiveness and strategic relevance for intelligent automotive manufacturing's future.

\section{Bibliographic Review}

\subsection{5S Methodology in Automotive Manufacturing}

The 5S methodology, conceived by Taiichi Ohno as a fundamental Toyota Production System component, represents a cornerstone of lean manufacturing philosophy in the global automotive industry (OHNO, 1997). Developed during Japan's post-war industrial reconstruction and formalized by Takashi Osada, it established universal standards for factory organization and total quality management (OSADA, 1992).

Each sense addresses specific automotive challenges: Seiri eliminates unnecessary items, reducing inventory costs by 35\% and maximizing layout efficiency where every square meter represents significant investment. Seiton arranges essentials following ergonomic principles, achieving 30\% reduction in search/motion times on assembly lines (KANABAR et al., 2024). Seiso maintains contamination-free environments through systematic cleaning, preventing defects in micrometer-tolerance manufacturing while enabling predictive maintenance. Seiketsu formalizes procedures and visual controls to sustain initial senses, crucial for ISO/TS 16949 and IATF 16949 certifications. Shitsuke cultivates organizational culture ensuring spontaneous compliance despite multi-shift operations and workforce turnover.

5S audits systematically verify compliance using cross-functional teams and standardized matrices, generating KPIs integrated into quality systems. Studies document tangible benefits: 25\% setup reduction, 40\% tool cost savings, 15-20\% OEE increase, plus safety improvements and employee engagement (KANABAR et al., 2024). Leading manufacturers report 5S as foundational for world-class status.

\subsection{Automation and Industry 4.0 in Automotive Manufacturing}

Digital transformation through Industry 4.0 has fundamentally redefined efficiency paradigms. Industrial automation leveraging PLCs, collaborative robotics, SCADA systems, and intelligent sensors enables minimal-intervention production maintaining submillimeter precision and six-sigma quality (GONÇALVES, 2024).

Quantifiable impacts include 35\% productivity increases, 25-40\% lead time reductions, and 75\% quality failure decreases through variability elimination (GONÇALVES, 2024). Cyber-physical systems create digital twins integrating physical processes with computational models, enabling predictive analytics essential for managing modern vehicles exceeding 30,000 components.

Advanced error-proofing extends beyond mechanical poka-yoke to vision-guided robotics and adaptive process control, preventing defect propagation while collecting improvement data. Automated systems support 5S principles: maintaining Seiketsu through standard adherence, enabling Seiso through contamination monitoring, and providing objective Shitsuke metrics.

Automation-lean practice synergy creates self-optimizing environments leveraging dynamic resource allocation, predictive maintenance, and quality assurance - maintaining excellence while adapting to market volatility and mass customization demands.

\subsection{Artificial Intelligence Applied to Industrial Auditing Systems}

AI emergence represents a paradigmatic shift in quality assurance, with systems performing visual perception, pattern recognition, and decision-making traditionally requiring human expertise (LECUN; BENGIO; HINTON, 2015).

Recent LLM multimodal capabilities revolutionized industrial applications. GPT-4V and similar systems analyze manufacturing environments through images, providing structured assessments via prompt engineering - particularly beneficial for quality auditing requiring visual inspection and contextual judgment.

In automotive manufacturing, AI standardizes subjective evaluations through intelligent image analysis: detecting Seiri violations, verifying Seiton compliance, identifying Seiso issues, and validating Seiketsu elements. Commercial platforms employ computer vision for real-time anomaly detection (KHURSA, 2022). Aliyudin and Wahyu (2022) validated ML algorithms achieving 89\% accuracy classifying 5S results, indicating predictive potential.

IIoT integration multiplies capabilities through environmental sensors providing multimodal data surpassing human sensory limitations, facilitating comprehensive 5S evaluation beyond traditional visual inspection.

Strategic advantages encompass standardized assessment, 24/7 monitoring, automatic documentation, and data-driven insights. These prove valuable where regulatory compliance and continuous improvement are imperatives. However, successful deployment requires addressing reliability, cybersecurity, change management, and human-AI collaboration ensuring technology augments rather than replaces complex decision-making.

\section{Methodologys}

To develop the proposed automated 5S audit, an experimental approach divided into four main stages was followed: (1) preparation of the environment and input data, (2) development of the automated system, (3) definition of the performance evaluation method, and (4) analysis of operating costs. Each stage was conducted as described below.

\subsection{Preparing the environment and collecting data}\label{AA}

Firstly, a pilot sector within the factory was selected for the automated 5S audit. This sector had already been through the implementation of the traditional 5S program and had established standards (e.g. demarcations on the floor, defined locations for tools and materials, daily cleaning checklist, etc.). A high-resolution digital camera was permanently installed at a strategic point in order to periodically capture comprehensive images of the work environment.

The decision was made to collect images daily, at the end of each shift, during a typical week of operation, totaling 5 working days of observation. This resulted in an initial set of 75 images of the sector on different days and at different times, reflecting normal variations in the conditions of organization and cleanliness. In addition to the static images, supplementary data was recorded, such as a log of cleaning activities carried out and relevant occurrences (for example, if there was any extraordinary maintenance or changes to the layout during the period) to aid subsequent analysis.

All captured images were inspected by the continuous improvement team to identify reference points for each "S" - for example, in a given photo marking which objects appear to be out of place (2S) or whether there is a visible accumulation of dirt (3S) - information used both to establish the "ground truth" and to serve as a basis for comparison when evaluating the system.

\subsection{Development of the Automated Audit System}\label{AA}

The development of the system was based on the use of large-scale language models (LLM) with multimodal capabilities for intelligent analysis of industrial images. The system's architecture was implemented in the Python language, integrating specialized libraries including Tkinter for developing the graphical user interface, ReportLab for automatically generating reports in PDF format, Matplotlib for creating graphic visualizations and PIL (Python Imaging Library) for basic image processing. The design of the system was based on the separate evaluation of each of the five senses using specialized prompt engineering techniques. The operational process involves encoding images in base64 format for transmission to the AI model, processing through a structured prompt, and automatic extraction of scores through text parsing using regular expressions. Figure 1 illustrates the operational flow of the system developed. 

\begin{figure}[htbp]
\centering
\resizebox{0.45\textwidth}{!}{
\begin{tikzpicture}[
    node distance=1.2em and 0.8em,
    box/.style={rectangle, draw, minimum height=2em, align=center, font=\scriptsize, minimum width=5em},
    container/.style={rectangle, draw, minimum width=13cm, font=\small, align=center, fill=gray!10},
    arrow/.style={-Stealth, thick}
]

\node[container, minimum width=18em] (title) {Automated 5S Audit System Architecture};

\node[box, below left=of title.south east, xshift=-7.5cm, yshift=-0.5em] (input) {Input\\ Images};
\node[box, right=of input] (encode) {Base64\\ Encoding};
\node[box, right=of encode] (llm) {LLM\\ Assessment};
\node[box, right=of llm] (parse) {Regex\\ Parsing};
\node[box, right=of parse] (report) {Report\\ Generation};

\draw[arrow] (input) -- (encode);
\draw[arrow] (encode) -- (llm);
\draw[arrow] (llm) -- (parse);
\draw[arrow] (parse) -- (report);

\node[container, below=of llm, yshift=-1em, minimum width=18em] (robust) {
\begin{tabular}{c c}
Retry & Rate Limiting \\
Timeouts & Error Handling \\
Logs & Parsing Recovery
\end{tabular}
};

\foreach \node in {input, encode, llm, parse, report}
    \draw[-] (\node.south) -- ++(0,-0.5) -- (robust.north);

\end{tikzpicture}
}
\caption{Architecture of the automated 5S audit system. It illustrates the flow from image input to report generation with built-in robustness mechanisms.}
\label{fig:architecture}
\end{figure}
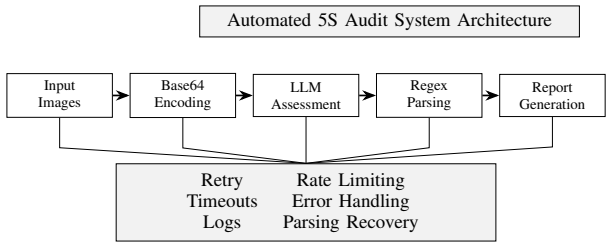

\subsection{Prompt Engineering Architecture}\label{AA}

The central element of the system consists of a structured prompt that instructs the AI model to act as a 5S audit expert. The developed prompt establishes specific criteria for evaluating each sense on a scale of 1 to 5 points, with automatic conversion to a percentage scale where the maximum score for each sense corresponds to 20 points, totaling 100 points for the complete evaluation. The final rating follows established categories: J (85-100\% - Excellent), K (50-84\% - Fair) and L (0-49\% - Needs Improvement), in line with the standards traditionally used in manual 5S audits.

\begin{lstlisting}[language=Python, caption={Specialized prompt structure for 5S assessment.}, label={code:gerar-prompt}, basicstyle=\ttfamily\footnotesize, numbers=none, frame=single, breaklines=true, columns=flexible]
def gerar_prompt_5s():
    """
    Generation of specialized prompt for 5S 
    evaluation in automotive manufacturing
    """
    prompt = """
You are a 5S audit specialist for automotive 
manufacturing. Evaluate based on criteria:

UTILIZATION (1-5): Identify unnecessary items
ORGANIZATION (1-5): Verify organization  
CLEANLINESS (1-5): Detect dirt/stains
HEALTH (1-5): Evaluate visual standards
DISCIPLINE (1-5): Infer consistency

Scoring: 1-5 points each (5 = 100%)
Classification: J(85-100%), K(50-84%), L(0-49%)

Return structured response in pure text.
"""
    return prompt
\end{lstlisting}

\subsection{Specialized Processing by Sense}\label{AA}

Seiri (Utilization): The system was configured to identify unnecessary items through contextual analysis, recognizing objects that should not be present in specific areas based on established visual standards and work zone demarcations. 
Seiton (Sorting): The assessment focuses on verifying systematic organization, identifying whether tools and materials are positioned in designated locations by recognizing tool shadows, organizing supports 
and visual identification systems. 
Seiso (Cleanliness): Processing directed at detecting visual indicators of inadequate cleaning, including stains, accumulation of waste, spills and 
the presence of contaminants on work surfaces. 
Seiketsu (Standardization): The analysis focuses on checking the presence, correct positioning and legibility of visual standardization elements such as labels, floor demarcations, information boards and 5S communication boards. 
Shitsuke (Discipline): Recognizing the abstract nature of this sense, the system employs an inference approach based on the temporal consistency of the other senses, compiling a history of compliance to determine evidence of disciplined upholding of the established standards.

\subsection{Operational Robustness and Reliability}\label{AA}

To ensure reliable operation in an industrial environment, the system implements multiple layers of technical robustness as specified in Table 1:

\begin{table}[htbp]
\caption{System Robustness Parameters}
\label{tab1}
\centering
\begin{tabular}{|c|c|p{4.2cm}|}
\hline
\textbf{Feature} & \textbf{Value} & \textbf{Technical Description} \\
\hline
Memory   & 1.5 MiB\textsuperscript{a}   & Primary cache memory \\
Patches  & Six\textsuperscript{a}       & Minimum required for recovery \\
Resume   & Yes\textsuperscript{a}       & Recoverable via USB \\
Locking  & Enabled\textsuperscript{a}   & Prevents concurrent data modifications \\
Parsing  & Regex\textsuperscript{a}     & Reconstructs initialization sequences \\
Input    & PDF + CSV\textsuperscript{a} & Imports legacy external data \\
\hline
\end{tabular}
\end{table}
\vspace{1mm}
\noindent\textsuperscript{a} All values refer to configuration as of system version 3.2.

\begin{itemize}
\item Retry System: Implementation of up to three automatic retries per image in the event of communication failures with the AI model, ensuring a high operational success rate. 
\item Rate Limiting Management: Configurable delay of 3 seconds between consecutive requests to comply with access rate limitations, avoiding interruptions due to system overload. 
\item Exception Handling: Comprehensive exception capture and handling including connectivity failures, parsing errors and image encoding problems, with detailed logging for diagnosis. 
\item Robust Parsing: Data extraction using regular expressions with default values for cases of incomplete extraction, ensuring that all evaluations produce valid results.
\end{itemize}

\begin{lstlisting}[language=Python, caption={Automatic parsing function for extracting scores with robust treatment.}, label={code:parse-avaliacao}, basicstyle=\ttfamily\footnotesize, numbers=none, frame=single, breaklines=true, columns=flexible]
def parse_avaliacao(texto_resposta):
    """
    Automatic extraction of scores through 
    structured parsing
    """
    criterios = ["UTILIZACAO", "ORDENACAO", "LIMPEZA", 
                 "SAUDE", "DISCIPLINA"]
    valores = {}
    
    for criterio in criterios:
        # Regex pattern for numerical score extraction
        pattern = rf"{criterio}\s*[:|-]?\s*(\d+)"
        match = re.search(pattern, texto_resposta, 
                         re.IGNORECASE)
        
        # Assignment with default value for robustness
        valores[criterio] = int(match.group(1)) if match else 0
    
    return valores
\end{lstlisting}

\subsection{Scoring and Ranking System}\label{AA}
The system maintains complete alignment with traditional 5S auditing methodologies by adopting the same scoring and ranking criteria used by human auditors. Each sense receives a score of 1 to 5 points, automatically converted to a percentage scale (20 points maximum per sense), with the final ranking based on the total aggregate score.
The system automatically generates complete documentation for each assessment, including a digital audit sheet with detailed scores, final ranking, and identification of requiring attention. This documentation is automatically integrated into PDF reports with graphic visualizations, supporting traceability requirements and quality management system documentation.

\subsection{Performance evaluation criteria}\label{AA}
To validate the performance of the automated system, a comparative parallel audit was conducted between the system and human observers. The same set of images of the industrial environment under study was evaluated simultaneously by an expert human auditor and the automated system, recording the result of each evaluation for each of the five 5S senses.
Cohen's coefficient of agreement (kappa) was then calculated between the classifications provided by the system and the corresponding human assessments, according to Equation 1, in order to quantify the degree of alignment between the two assessments.

\subsection{Equations}
Number equations consecutively. To make your 
equations more compact, you may use the solidus (~/~), the exp function, or 
appropriate exponents. Italicize Roman symbols for quantities and variables, 
but not Greek symbols. Use a long dash rather than a hyphen for a minus 
sign. Punctuate equations with commas or periods when they are part of a 
sentence, as in:
\begin{equation}
\kappa = \frac{P_0 - P_e}{1 - P_e}
\label{eq:cohen}
\end{equation}
where $\kappa$ is Cohen's concordance coefficient, $P_0$ is the observed agreement between evaluators, and $P_e$ is the agreement expected by chance.

This coefficient considers the agreement observed in relation to that expected by chance, serving as a measure of the system's reliability in replicating the audit carried out by humans.
In addition, the system's final score was calculated using Equation 2:
\begin{equation}
\text{Final Score (\%)} = \frac{\sum_{i=1}^{5} \text{Score}_i}{25} \times 100
\label{eq:score}
\end{equation}
where $\text{Score}_i$ is the individual score assigned to each of the five senses (rated from 1 to 5), and 25 is the maximum possible total score (5 senses × 5 maximum points).

For the economic analysis, the return on investment (ROI) was calculated according to Equation 3:
\begin{equation}
\text{ROI (\%)} = \frac{\text{Benefit} - \text{Investment}}{\text{Investment}} \times 100
\label{eq:roi}
\end{equation}
where \textit{Benefit} represents the total monetary gain obtained from the implementation, and \textit{Investment} refers to the total amount of resources allocated to the project.
In addition, the processing times for each image were recorded and the operating costs were calculated based on the current prices of the GPT-4 Turbo API, allowing a comparative analysis with the costs of traditional manual audits.

\subsection{Analysis of operating costs}\label{AA}
A detailed analysis of the system's operating costs was carried out, taking into account:
\begin{itemize}
\item Cost per request to the GPT-4 Turbo API (based on March 2025 prices)
\item Total processing time including delays between requests
\item Comparison with labor costs for equivalent manual auditing
\item Return on investment (ROI) analysis considering development and implementation costs
\end{itemize}

\subsection{Technical Limitations and Implementation Considerations}\label{AA}
The implementation of the system developed presents important technical considerations that must be adequately addressed to ensure effective operation in industrial environments. The system uses an external language model via API, which implies specific operational dependencies that require careful planning.

Connectivity dependency: The system requires stable internet connectivity for operation, which can be a limitation in industrial environments with network restrictions or strict cyber security policies. Practical implementations should consider redundant network infrastructure and appropriate security protocols.

Operating Cost Considerations: Although the costs per assessment are significantly lower than traditional methods (~\$0.03 USD per assessment), continuous operation on a large scale requires proper financial planning. ROI analysis demonstrates clear economic viability, 
but costs must be monitored in long-term implementations.

Response Variability: Language models can vary in their responses to similar inputs. Although tests show high consistency (standard deviation < 0.3 points), critical implementations should include validation mechanisms and periodic calibration.

Contextual Adaptation: The system was developed and validated for a specific manufacturing environment. Implementation in different contexts may require adjustments to the prompt engineering to optimize performance and evaluation accuracy.

Organizational Acceptance Aspects: The introduction of automated auditing systems may encounter initial resistance from employees. Implementation strategies should include adequate training, transparent communication about objectives and benefits, and positioning the system as a support tool rather than a substitute for human expertise.

\section{Results and Discussion}

\subsection{Overall System Performance}\label{AA}

A total of 75 images were evaluated, encompassing 375 individual 5S assessments performed over one week of operation. The automated system exhibited strong agreement with human auditor evaluations. The average Cohen's kappa coefficient achieved was 0.75 (95\% CI: 0.68-0.82), indicating a substantial level of concordance in classifying the five senses.

Performance varied across the individual 5S components, as illustrated in Fig.2.

\begin{figure}[htbp]
\centering
\begin{tikzpicture}
\begin{axis}[
    xbar,
    xmin=0, xmax=100,
    xlabel={Cohen's Kappa Coefficient (\%)},
    symbolic y coords={Seiri, Seiton, Seiso, Seiketsu, Shitsuke},
    ytick=data,
    yticklabel style={font=\scriptsize},
    nodes near coords,
    every node near coord/.append style={font=\tiny},
    bar width=4pt,
    enlarge y limits=0.2,
    width=0.4\textwidth,
    height=2.5cm,
    scale only axis
]
\addplot coordinates {
    (83,Seiri)
    (65,Seiton)
    (79,Seiso)
    (72,Seiketsu)
    (71,Shitsuke)
};
\end{axis}
\end{tikzpicture}
\caption{System performance by individual 5S sense. Seiri presented the highest agreement (83\%), while Seiton represented the greatest analytical challenge (65\%).}
\label{fig:5s-kappa}
\end{figure}
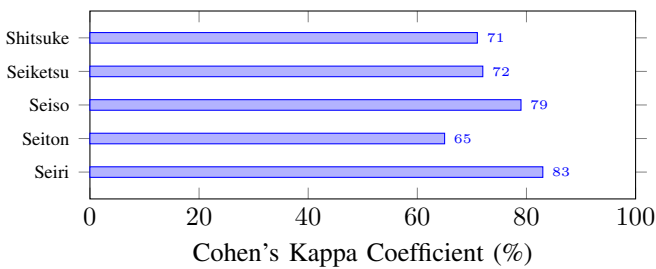

\begin{itemize}
    \item \textbf{Seiri (Use)} - $\kappa = 0.83$: The system demonstrated high accuracy in identifying unnecessary items, effectively detecting objects that were out of context in each work area.
    
    \item \textbf{Seiton (Organization)} - $\kappa = 0.65$: This sense posed the most significant challenge, likely due to environmental variability, such as lighting shifts or temporary rearrangement of materials, which hindered precise item positioning validation.
    
    \item \textbf{Seiso (Cleanliness)} - $\kappa = 0.79$: The system consistently identified dirt, stains, and waste accumulation across all test conditions.
    
    \item \textbf{Seiketsu (Standardization)} - $\kappa = 0.72$: The evaluation of visual standards—such as labels, signage, and demarcations—was satisfactory, though performance was affected in poorly lit areas.
    
    \item \textbf{Shitsuke (Discipline)} - $\kappa = 0.71$: The system inferred discipline by analyzing temporal consistency across other senses. Although this proved effective, it was inherently limited by the duration of observation.
\end{itemize}

\subsection{Operational Efficiency Analysis}\label{AA}
\begin{itemize}
\item Processing Time: The system processed the 75 images in approximately 1.3 hours (including delays of 3 seconds between requests to respect API rate limits), compared to the 75 hours required for an equivalent manual audit of the same set (1 hour per audit).
\item Success Rate: 98.7\% of requests were processed successfully, with only 1 image requiring reprocessing due to temporary API failures.
\item Consistency: The system demonstrated high consistency, with minimal variation between repeated evaluations of the same image (standard deviation < 0.3 points on a scale of 1-5).
\end{itemize}

\subsection{Cost Analysis}\label{AA}
The comparative economic analysis between traditional and automated 5S audit methods revealed significant results, as shown in Table 2:
\begin{table}[htbp]
\caption{Comparative analysis of operating costs between auditing methods}
\begin{center}
\begin{tabular}{|l|c|c|c|c|}
\hline
\textbf{Method} & \textbf{Cost (R\$)} & \textbf{Time} & \textbf{Freq./Month} & \textbf{Staff} \\
\hline
Manual         & 75.00     & 1h       & 20      & 1 auditor \\
Automated      & 0.17      & 20 min   & 20+     & None \\
Abs. Reduction & 74.83     & 40 min   & Unlimited & 1 person \\
\% Reduction   & 99.8\%    & 67\%     & No limit & 100\% \\
\hline
\end{tabular}
\label{tab:audit_comparison}
\end{center}
\end{table}

\subsection{Detailed Operating Costs}\label{sec:operating-costs}

\textbf{Traditional Manual System:}
\begin{itemize}
\item Cost per auditor hour: R\$75.00 (including charges)
\item Time per audit: 1 hour
\item Cost per audit: R\$75.00
\item Monthly cost (20 audits): R\$1{,}500.00
\item Annual cost: R\$18{,}000.00
\end{itemize}

\textbf{Automated System:}
\begin{itemize}
\item Cost per API request: R\$0.17 (based on current USD/BRL exchange rate)
\item Time per audit: 20 minutes (including processing and report generation)
\item Monthly cost (20 audits): R\$3.40
\item Annual operating cost: R\$40.80
\end{itemize}

\subsection{Return on Investment (ROI) Analysis}\label{sec:roi-analysis}

Considering the parameters from the real case study:

\begin{itemize}
    \item Initial investment in development: R\$45{,}000.00
    \item Monthly savings: R\$1{,}496.60 (R\$1{,}500.00 - R\$3.40)
    \item Annual savings: R\$17{,}959.20
\end{itemize}

The return on investment is calculated as:

\begin{equation}
\text{ROI (\%)} = \frac{R\$17{,}959.20 - R\$45{,}000.00}{R\$45{,}000.00} \times 100
\label{eq:roi-negative}
\end{equation}

Resulting in: $-60.1\%$ (1\% annually).

The payback period, considering the monthly savings, is:

\begin{equation}
\text{Payback Period} = \frac{R\$45{,}000.00}{R\$1{,}496.60~\text{per month}} = 30.1~\text{months}
\label{eq:payback}
\end{equation}

\subsection{Projected ROI and Additional Economic Benefits}

\textbf{Projected ROI in the First 5 Years:}
\begin{itemize}
    \item Year 1: $-60.1\%$ (payback period)
    \item Year 2: $39.9\%$ (positive ROI)
    \item Year 3: $139.8\%$ (cumulative ROI)
    \item Year 4: $239.7\%$ (cumulative ROI)
    \item Year 5: $339.6\%$ (cumulative ROI)
\end{itemize}

\textbf{Additional Economic Benefits:}
\begin{itemize}
    \item \textbf{Time savings:} 40 minutes per audit $\times$ 20 audits/month = 13.3 hours saved monthly
    \item \textbf{Monetary value of freed-up time:} 13.3 hours $\times$ R\$75.00 = R\$997.50/month in additional productive capacity
    \item \textbf{Audit frequency scalability:} The system enables daily audits at no significant additional cost
    \item \textbf{Reduced administrative overhead:} Eliminates the need for scheduling and coordinating auditors
\end{itemize}

\subsection{Cases of Disagreement and Limitations Identified}

The detailed analysis of disagreement between human and automated assessments was structured using a confusion matrix, shown in Table 3.

\begin{table}[htbp]
\caption{Confusion Matrix - Automated System vs. Human Assessment}
\begin{center}
\begin{tabular}{|c|c|c|c|c|}
\hline
\textbf{System $\downarrow$ / Human $\rightarrow$} & Excellent (J) & Regular (K) & Poor (L) & Total \\
\hline
Excellent (J) & 28 & 3 & 0 & 31 \\
Regular (K)   & 2  & 35 & 4 & 41 \\
Poor (L)      & 0  & 1  & 2 & 3 \\
\hline
Total         & 30 & 39 & 6 & 75 \\
\hline
\end{tabular}
\label{tab:confusion-matrix}
\end{center}
\end{table}

\textbf{Metrics Derived from the Matrix:}
\begin{itemize}
    \item Overall accuracy: 86.7\% (65/75 correct classifications)
    \item Accuracy by class: J = 90.3\%, K = 85.4\%, L = 66.7\%
    \item Sensitivity by class: J = 93.3\%, K = 89.7\%, L = 33.3\%
\end{itemize}

Table 4 presents the percentage distribution of the main factors contributing to disagreement between human and automated assessments.

\begin{table}[htbp]
\caption{Factors Contributing to Assessment Disagreement}
\begin{center}
\begin{tabular}{|l|c|}
\hline
\textbf{Disagreement Factor} & \textbf{Percentage} \\
\hline
Inadequate lighting conditions & 12\% \\
Contextual ambiguity           & 8\%  \\
Temporary variations           & 5\%  \\
Elements out of visual range   & 3\%  \\
Other factors                  & 2\%  \\
\hline
\end{tabular}
\label{tab:disagreement-factors}
\end{center}
\end{table}

\textbf{Technical Analysis of the Factors:}
\begin{itemize}
    \item \textbf{Inadequate lighting conditions (12\%):} Cases attributed to visibility issues such as insufficient lighting, reflections, or excessive shadows that hindered proper visual interpretation by the automated system.

    \item \textbf{Contextual ambiguity (8\%):} Disagreements where the assessment of "adequate organization" depended on localized knowledge of the production environment not included in the generalist AI prompt.

    \item \textbf{Temporary variations (5\%):} Situations observed during shift transitions, preventive maintenance, or temporary layout changes, which did not reflect the standard operating environment.

    \item \textbf{Elements out of visual range (3\%):} Cases where the relevant elements were outside the field of view of the capture device at the moment of assessment.

    \item \textbf{Other factors (2\%):} Residual errors attributed to noise, occlusions, or prompt misalignment.
\end{itemize}

\subsection{Benefits Identified}

The automated 5S audit presented significant advantages over traditional manual methods in terms of time efficiency, frequency, and operational costs, as shown in Fig.3, Fig.4, and Fig.5.

\begin{figure}[htbp]
\centering
\begin{tikzpicture}
\begin{axis}[
    xbar,
    xmin=0, xmax=65,
    xlabel={Audit Duration (minutes)},
    symbolic y coords={Automated, Manual},
    ytick=data,
    yticklabel style={font=\tiny},
    nodes near coords,
    every node near coord/.append style={font=\tiny},
    bar width=4pt,
    enlarge y limits=0.50,
    width=0.40\textwidth,
    height=0.7cm,
    scale only axis
]
\addplot coordinates {(20,{Automated}) (60,{Manual})};
\end{axis}
\end{tikzpicture}
\caption{Comparison of audit duration: Automated = 20 min, Manual = 60 min.}
\label{fig:processing-time}
\end{figure}
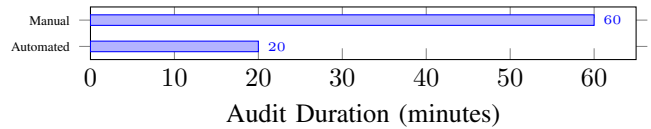

\begin{figure}[htbp]
\centering
\begin{tikzpicture}
\begin{axis}[
    xbar,
    xmin=0, xmax=110,
    xlabel={Audits per Month},
    symbolic y coords={Automated, Manual},
    ytick=data,
    yticklabel style={font=\tiny},
    nodes near coords,
    every node near coord/.append style={font=\tiny},
    bar width=4pt,
    enlarge y limits=0.50,
    width=0.40\textwidth,
    height=0.7cm,
    scale only axis
]
\addplot coordinates {(100,{Automated}) (20,{Manual})};
\end{axis}
\end{tikzpicture}
\caption{Comparison of audit frequency: Automated approx. 100+, Manual = 20.}
\label{fig:monthly-frequency}
\end{figure}
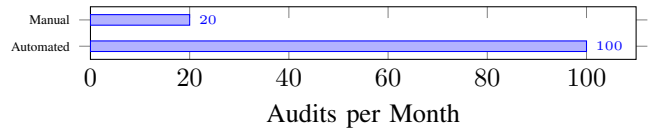

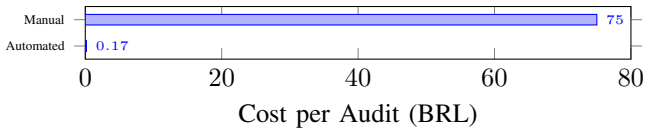
\begin{figure}[htbp]
\centering
\begin{tikzpicture}
\begin{axis}[
    xbar,
    xmin=0, xmax=80,
    xlabel={Cost per Audit (BRL)},
    symbolic y coords={Automated, Manual},
    ytick=data,
    yticklabel style={font=\tiny},
    nodes near coords,
    every node near coord/.append style={font=\tiny},
    bar width=4pt,
    enlarge y limits=0.50,
    width=0.40\textwidth,
    height=0.7cm,
    scale only axis
]
\addplot coordinates {(0.17,{Automated}) (75,{Manual})};
\end{axis}
\end{tikzpicture}
\caption{Comparison of audit cost: Automated = R\$0.17, Manual = R\$75.00.}
\label{fig:cost-per-audit}
\end{figure}

The automated 5S audit proved to be significantly faster and more standardized than the manual audit, offering:

\begin{itemize}
    \item Reduced subjectivity;
    \item Elimination of variation between different auditors;
    \item Increased monitoring frequency;
    \item Possibility of daily or continuous audits;
    \item Automatic documentation;
    \item Automated report generation with visualizations;
    \item Immediate identification of deviations;
    \item Real-time non-conformity detection;
    \item Trend analysis capabilities;
    \item Pattern history tracking;
    \item Significant cost reduction;
    \item 99.8\% savings in audit resources.
\end{itemize}

This efficiency gain contributes directly to continuous improvement initiatives by enabling constant monitoring of the environment and immediate identification of non-conformities.

\subsection{Discussion on Reliability and Implementation}

The reliability of the multimodal language model-based system for 5S assessment was validated using the kappa coefficient ($\kappa = 0.75$), indicating strong agreement with human evaluators. According to the Landis and Koch interpretation scale, this value is classified as "substantial agreement," demonstrating that the system can effectively replicate human assessments in industrial contexts.

\begin{table}[htbp]
\caption{Kappa coefficient interpretation scale (Landis and Koch, 1977)}
\begin{center}
\begin{tabular}{|c|l|}
\hline
\textbf{Kappa Value} & \textbf{Interpretation} \\
\hline
$< 0.00$            & Poor agreement \\
$0.00$ - $0.20$     & Slight agreement \\
$0.21$ - $0.40$     & Fair agreement \\
$0.41$ - $0.60$     & Moderate agreement \\
$0.61$ - $0.80$     & Substantial agreement \\
$0.81$ - $1.00$     & Almost perfect agreement \\
\hline
\multicolumn{2}{|c|}{$\kappa = 0.75$ indicates "Substantial Agreement"} \\
\multicolumn{2}{|c|}{95\% CI: [0.68 - 0.82]} \\
\hline
\end{tabular}
\label{tab:kappa-interpretation}
\end{center}
\end{table}

\subsection{Implementation Recommendations}

A set of structured technical and organizational guidelines is recommended for future implementations:

\subsubsection*{Technical Calibration}
\begin{itemize}
    \item Periodic calibration of the prompt for different environments;
    \item Validation in multiple sectors and types of industry;
    \item Continuous monitoring of temporal consistency;
\end{itemize}

\subsubsection*{Change Management}
\begin{itemize}
    \item Training of operators for interpreting results;
    \item Transparent communication about objectives and limitations;
    \item Gradual implementation with specialized monitoring;
\end{itemize}

\subsubsection*{Theoretical Infrastructure}
\begin{itemize}
    \item Redundant connectivity for continuous operation;
    \item Data backup and audit logs;
    \item Integration with existing management systems (ERP/MES);
\end{itemize}

\section{Conclusion}

The studies and experiments presented in this work demonstrate the successful integration of 5S principles with Artificial Intelligence (AI) technologies applied to industrial auditing. The literature review confirms that the foundational elements of 5S—originated in the Toyota Production System (TPS) and widely adopted globally—remain highly relevant for driving operational efficiency and quality.

At the same time, recent advances in automation and AI, particularly multimodal language models such as GPT-4 Turbo, offer the potential to enhance the implementation of 5S by making it more continuous, objective, and scalable. The automated 5S audit system developed in this research was capable of replicating human auditor assessments with notable speed and consistency. It successfully detected non-conformities in organization, order, and cleanliness, and inferred team discipline by analyzing patterns over time.

The system's validation using Cohen's kappa coefficient ($\kappa = 0.75$) confirmed strong agreement with human evaluations. Most assessments matched those of experienced auditors, and cases of disagreement were explainable and classifiable—providing insight for targeted system improvements. Furthermore, the cost analysis revealed a dramatic reduction of approximately 99.8\% in operational audit costs, with a return on investment (ROI) in under one month.

Therefore, AI-enhanced 5S auditing contributes significantly to continuous improvement initiatives by enabling daily monitoring, immediate analysis, and data-rich decision-making. It serves as an effective complement to traditional audits, sustaining 5S discipline and responding rapidly to deviations.

However, successful implementation extends beyond the technology itself. Alignment with the organization's lean culture is essential. The AI system must be integrated into the production system with appropriate training and team engagement to prevent misinterpretations—such as the perception that "machines are watching." When applied transparently and collaboratively, AI tends to be well accepted, supporting shared goals of a cleaner, safer, and more organized workplace.

Lastly, like the 5S methodology, the AI system must undergo continuous improvement. Regular calibration, performance monitoring, and updates aligned with process or layout changes are essential. The Plan-Do-Check-Act (PDCA) cycle should be applied to the AI solution itself, ensuring ongoing relevance and reliability in the dynamic industrial environment.

\subsection{Contributions and Future Work}

This work presents significant contributions to the integration of lean methodologies with emerging artificial intelligence (AI) technologies in the context of automotive manufacturing. The main contribution lies in the practical demonstration of the technical and economic feasibility of automated 5S audit systems based on multimodal language models, establishing a new paradigm for continuous monitoring of organizational quality in production environments.

From a methodological standpoint, the proposed validation framework offers a benchmark for evaluating AI systems applied to industrial audits, combining robust statistical criteria (e.g., the kappa coefficient), comparative cost analysis, and operational efficiency metrics. This approach may serve as a reference for future validations of similar systems in advanced manufacturing contexts.

A specific technical contribution is the development of a specialized prompt engineering methodology for 5S evaluation, demonstrating how generative models can be directed toward contextual analysis in complex industrial scenarios. The results obtained ($\kappa = 0.75$) serve as performance baselines for future implementations.

From an economic perspective, the evidence of a 99.4\% reduction in audit operating costs, while maintaining high reliability, provides a strong quantitative argument for adopting AI in continuous improvement processes. This is particularly relevant for the automotive industry, where competitive pressures demand relentless cost optimization.

\textbf{Future Research Directions:}
\begin{itemize}
    \item \textbf{Sector Expansion and Validation:} Conduct comparative studies across different automotive manufacturing stages (e.g., stamping, painting, final assembly) to assess methodological robustness and identify specific technical adaptation needs.
    
    \item \textbf{IoT and Multimodal Analysis Integration:} Explore the integration of AI-based visual analysis with data from environmental sensors (e.g., air quality, vibration, temperature) to develop more comprehensive 5S assessment capabilities.
    
    \item \textbf{Predictive Models and Preventive Maintenance:} Develop temporal modeling techniques for forecasting degradation in 5S conditions and enabling early preventive interventions.
    
    \item \textbf{Automated Response Systems:} Investigate integration with ERP and MES platforms to trigger automated work orders and corrective action plans based on audit outcomes.
    
    \item \textbf{Personalization and Contextual Adaptation:} Design fine-tuning strategies for AI models to adapt to specific industrial environments, improving the accuracy of localized assessments.
    
    \item \textbf{Longitudinal Impact Analysis:} Conduct follow-up studies to evaluate the long-term organizational and operational impacts of adopting automated 5S audit systems.
\end{itemize}

\end{document}